\documentclass[conference]{IEEEtran}
\IEEEoverridecommandlockouts
\usepackage[noadjust]{cite}
\usepackage{amsmath,amssymb,amsfonts}
\usepackage{algorithmic}
\usepackage{graphicx}
\usepackage{textcomp}
\usepackage{xcolor}

\usepackage{subcaption}
\usepackage{caption}
\DeclareCaptionJustification{justified}{\justifying}
\usepackage[raggedrightboxes]{ragged2e}
\usepackage{algorithm}
\usepackage{algorithmic}
\usepackage{float}
\usepackage{multirow}
\usepackage{booktabs} 
\usepackage{tabularx}
\usepackage{multirow}
\usepackage{threeparttable}
\usepackage{bbding}
\usepackage{pifont}
\usepackage{wasysym}
\usepackage{amssymb}
\usepackage{makecell}
\usepackage{arydshln}
\usepackage[colorlinks=true, citecolor=blue]{hyperref}
\def\BibTeX{{\rm B\kern-.05em{\sc i\kern-.025em b}\kern-.08em
    T\kern-.1667em\lower.7ex\hbox{E}\kern-.125emX}}
\begin{document}

\title{Compressed Video Quality Enhancement: Classifying and Benchmarking over Standards}

\author{
\IEEEauthorblockN{Xiem HoangVan\textsuperscript{1}, Dang BuiDinh\textsuperscript{1}, Sang NguyenQuang\textsuperscript{2}, Wen-Hsiao Peng\textsuperscript{2}}
\IEEEauthorblockA{\textsuperscript{1}\textit{Faculty of Electronics and Telecommunications, VNU University of Engineering and Technology, Hanoi, Vietnam}}
\IEEEauthorblockA{\textsuperscript{2}\textit{Department of Computer Science, National Yang Ming Chiao Tung University, Hsinchu, Taiwan}}
}
\maketitle

\begin{abstract}
Compressed video quality enhancement (CVQE) is crucial for improving user experience with lossy video codecs like H.264/AVC, H.265/HEVC, and H.266/VVC. While deep learning based CVQE has driven significant progress, existing surveys still suffer from limitations: lack of systematic classification linking methods to specific standards and artifacts, insufficient comparative analysis of architectural paradigms across coding types, and underdeveloped benchmarking practices. To address these gaps, this paper presents three key contributions. First, it introduces a novel taxonomy classifying CVQE methods across architectural paradigms, coding standards, and compressed-domain feature utilization. Second, it proposes a unified benchmarking framework integrating modern compression protocols and standard test sequences for fair multi-criteria evaluation. Third, it provides a systematic analysis of the critical trade-offs between reconstruction performance and computational complexity observed in state-of-the-art methods and highlighting promising directions for future research. This comprehensive review aims to establish a foundation for consistent assessment and informed model selection in CVQE research and deployment.
\end{abstract}

\section{Introduction}
\label{sec:introduction}
\begin{figure}[!t]
    \centering
    \centerline{\includegraphics[width=0.5\textwidth]{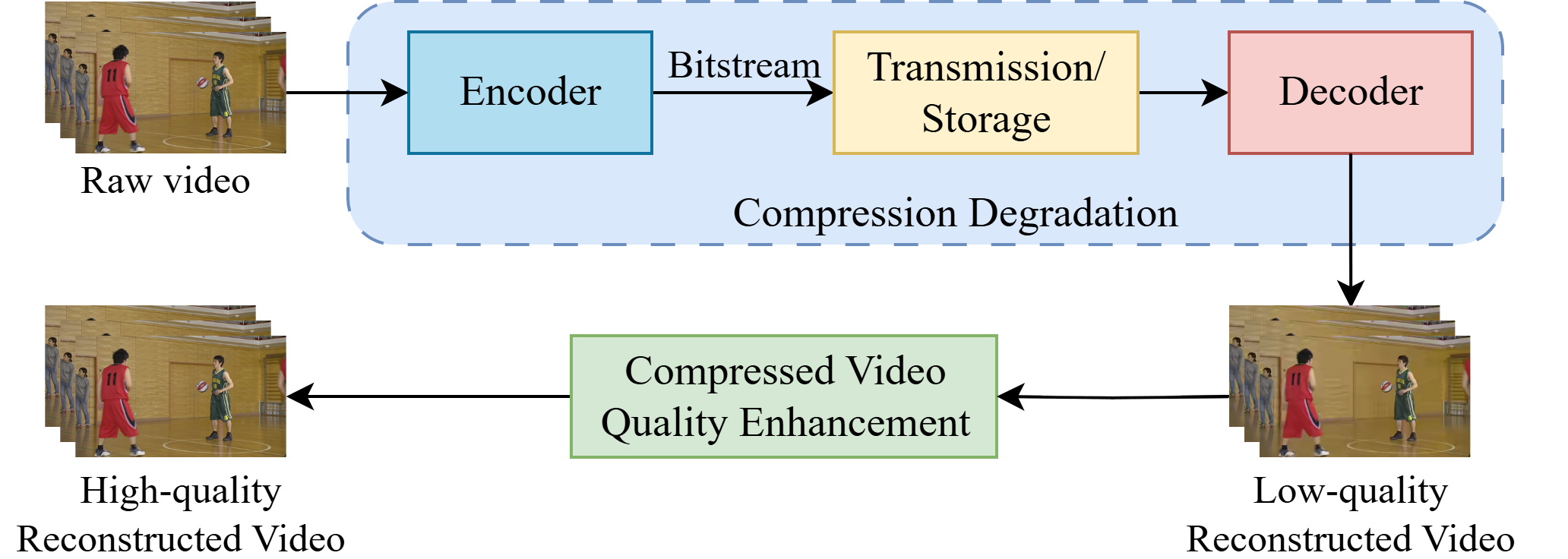}}
    \caption{Schematic illustration of end-to-end Compressed Video Quality Enhancement (CVQE) Flow.}
    \vspace{-0.6cm}
    \label{fig:flow-diagram}
\end{figure}

Video coding has been extensively researched and developed to compress video data while preserving optimal quality for end users, including standards such as H.264/AVC~\cite{avc-overview}, H.265/HEVC~\cite{hevc-overview}, and H.266/VVC~\cite{vvc-overview}. However, since these video codecs typically use a lossy compression mechanism, the quality of the reconstructed video is inevitably degraded. To address this limitation, various methods for compressed video quality enhancement (CVQE) have been introduced~\cite{ovqe, stdf, stff, cpga,mdeformer,tvqe,caef, ctve}. CVQE models can be categorized into single-frame-based methods~\cite{qe-cnn,sefcnn,gl-fusion,irnn}, which take a single frame as input, and multi-frame-based methods~\cite{ovqe,stff,cpga,rivulet,mfqe}, which leverage multiple consecutive frames to exploit spatiotemporal correlations. As shown in Fig.~\ref{fig:flow-diagram}, CVQE models are employed to reconstruct high-quality videos from their low-quality counterparts obtained after the video compression and decoding process.

Although Yue~\emph{et al.}~\cite{survey} has recently conducted a survey of CVQE techniques, notable gaps remain. First, existing reviews lack a systematic framework linking enhancement methods to specific video coding standards and their associated artifacts. Second, architectural comparisons across coding types (intra- vs. inter-frame) are limited, especially for emerging standards like VVC. Third, benchmarking is underdeveloped, with few studies providing unified evaluations of perceptual quality and computational complexity under standardized conditions. These gaps hinder consistent performance assessment and practical model selection for application-specific deployment.

To address these limitations, this work conducts a comprehensive survey of CVQE over video coding standards with three key contributions:
\begin{itemize}
    \item A novel taxonomy that classifies CVQE methods across architectural paradigms, coding standards, and compressed-domain feature utilization
    \item A unified benchmarking framework that integrates modern video compression protocols and standard test sequences, enabling direct comparisons using multi-criteria evaluation metrics.
    \item A systematic analysis of the trade-offs between reconstruction performance and computational complexity, highlighting critical inverse relationships observed across state-of-the-art CVQE methods.
\end{itemize}

To demonstrate this work, the rest of this paper is organized as follows. Section~\ref{sec:classification} presents a structured classification of CVQE approaches across standards, covering model architectures, codec-specific adaptations, and compressed-domain information utilization. Meanwhile, Section~\ref{sec:benchmarking} details benchmarking methodologies including common test sequences, parameters,and evaluation metrics, and further analyzes performance trends for state-of-the-art CVQE methods, including computational efficiency comparisons. Finally, Section~\ref{sec:conclusions} concludes the study by synthesizing key insights and implications for the field.

\section{CVQE over Standards: A Classification}
\label{sec:classification}
This section provides a structured classification of recent approaches to CVQE, focusing on three key perspectives: model architectures, codec-specific adaptations, and the utilization of compressed-domain information.
\begin{table*}[!t]
\centering
\scriptsize
\caption{Compressed video quality enhancement methods categorized by video codec and network architecture.}
\label{tab:overview_table}
\newcolumntype{R}[1]{>{\raggedleft\arraybackslash}p{#1}}  
\newcolumntype{C}[1]{>{\Centering\arraybackslash}p{#1}}
\renewcommand{\arraystretch}{1.2} 


\begin{tabular}{@{} C{0.05\textwidth} @{} C{0.06\textwidth} @{} C{0.04\textwidth} @{} C{0.06\textwidth} @{} C{0.08\textwidth} @{} C{0.7\textwidth} @{}}
\hline
\multicolumn{2}{c}{Category}                               & CNN & Attention & \makecell{Others \\ \scriptsize{(e.g., MLP)}} & Representive Methods                                                                                                                                                                        \\ 
\hline
\multirow{3}{*}{\makecell{H.264/\\AVC}}  & Intra                  & \checkmark                       &           &                            & QE-CNN~\cite{qe-cnn}                                                                                                                                                                                      \\ 
\cline{2-6}
                            & \multirow{2}{*}{Inter} & \checkmark                       &           &                            & CAEF~\cite{caef}                                                                                                                                                                                        \\ 
\cline{3-6}
                            &                              & \checkmark                       & \checkmark         &                            & MetaBit~\cite{metabit}                                                                                                                                                                                     \\ 
\cline{1-6}
\multirow{8}{*}{\makecell{H.265/\\HEVC}} & \multirow{1}{*}{Intra}                 & \multirow{1}{*}{\checkmark}                       &           &                            & SEFCNN~\cite{sefcnn}, GL-Fusion~\cite{gl-fusion},                                                                                                                                                RNAN~\cite{rnan}, DnCNN~\cite{dncnn}, VRCNN~\cite{vrcnn},                                                                                                                                        DCAD~\cite{dcad}, AR-CNN~\cite{arcnn}                                                                                                                                                                                                    \\ 
\cline{2-6}
                            & \multirow{7}{*}{Inter} & \multirow{4}{*}{\checkmark}                       &           &                            & MFMN~\cite{mfmn}, STDF-DSCS~\cite{stdf-dscs}, STLVQE~\cite{stlvqe}, EMAFA~\cite{emafa}, TGAFNet~\cite{tgafnet}, BQEV~\cite{bqev}, MFQR~\cite{mfqr}, TWIT~\cite{twit}, STDN~\cite{stdn}, Fast-MFQE~\cite{fast-mfqe}, STDR~\cite{stdr}, PIMnet~\cite{pimnet}, VIG~\cite{vig}, FastCNN~\cite{fastcnn}, MRDN~\cite{mrdn}, CF-STIF~\cite{cf-stif}, DCNGAN~\cite{dcngan}, CLSTA~\cite{clsta}, RFDA~\cite{rfda}, STDF-R3L~\cite{stdf}, BSTN~\cite{bstn}, MW-GAN~\cite{mw-gan}, MFQEv2~\cite{mfqe2.0}, MFQE~\cite{mfqe}, MGANet~\cite{mganet} \\ \cline{3-4} 
\cline{3-6}
                            &                              & \multirow{1}{*}{\checkmark}                       & \multirow{1}{*}{\checkmark}         &                            & IIRNet~\cite{iirnet}, CPGA~\cite{cpga}, SVKAM~\cite{svkam}, STA~\cite{sta}, OVQE~\cite{ovqe}, STAWFN~\cite{stawfn}, STIB~\cite{stib}             \\ 
\cline{3-6}
                            &                              & \multirow{2}{*}{\checkmark}                        & \multirow{2}{*}{\checkmark}          & \multirow{2}{*}{\checkmark}                          & RivuletMLP~\cite{rivulet}, MSCAA~\cite{mscaa}, STFF~\cite{stff}, NU-CLass Net~\cite{nu-class-net}, MDEformer~\cite{mdeformer}, TVQE~\cite{tvqe}, CTVE~\cite{ctve}, M-Swin~\cite{m-swin}, PixRevive~\cite{pixrevive}, STCF~\cite{stcf}, PeQuENet~\cite{pequenet}, EAAGA~\cite{eaaga} \\ 
\hline
\multirow{3}{*}{\makecell{H.266/\\VVC}}  & Intra                  & \checkmark                       &           &                            & IRNN~\cite{irnn}, ILF-QE~\cite{ilf-qe},                                             HGRDN~\cite{hgrdn} \\ 
\cline{2-6}
                            & \multirow{2}{*}{Inter} & \checkmark                       &           &                            & STENet~\cite{stenet}, MFQE\_VVC~\cite{mfqe-vvc} \\ 
\cline{3-6}
                            &                              & \checkmark                       & \checkmark         &                            & OVQE\_VVC~\cite{ovqe-vvc} \\
\hline         
\end{tabular}
\vspace{-0.5cm}
\end{table*}

\subsection{CVQE over model architectures}
\begin{figure}[!t]
    \centering
    \centerline{\includegraphics[width=0.45\textwidth]{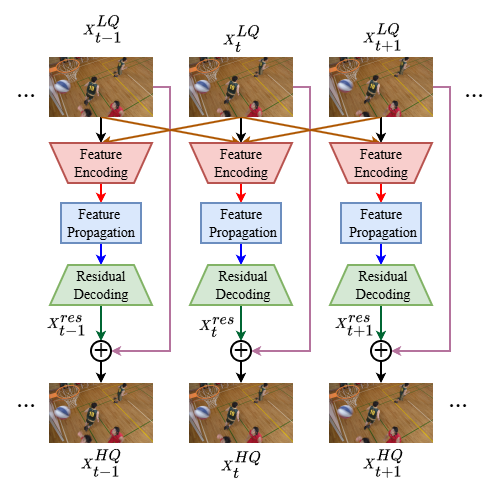}}
    \caption{Overall architecture of CVQE methods.}
    \label{fig:overall}
\vspace{-0.5cm}
\end{figure}
In overall, the architecture of CVQE models generally comprises three principal components: Feature Encoding (FE), Feature Propagation (FP), and Residual Decoding (RD), as illustrated in Fig.~\ref{fig:overall} and formalized in Eq.~\ref{eq:cvqe}:
\begin{equation}
\label{eq:cvqe}
X_{t}^{res} = RD\left ( FP\left ( FE\left (X_{t-m}^{LQ},...,X_{t}^{LQ},...,X_{t+m}^{LQ} \right ) \right ) \right ).
\end{equation}
Specifically, to enhance the quality of a low-quality reconstructed frame $X_{t}^{LQ}$, single-frame CVQE models take it as input, whereas multi-frame methods incorporate adjacent frames as temporal references, denoted by $\left\{X_{t-m}^{LQ}, \ldots, X_{t}^{LQ}, \ldots, X_{t+m}^{LQ} \right\}$. The process begins with the FE module, which extracts shallow features from these inputs. The resulting features are then propagated via the FP module and subsequently refined by the RD module to yield a fused representation. Finally, this fused feature map is decoded to produce an enhanced residual $X_{t}^{res}$, which is added to the original input to reconstruct the high-quality frame $X_{t}^{HQ}$: 
\begin{equation}
\label{eq:hq_frame}
X_{t}^{HQ} = X_{t}^{LQ} + X_{t}^{res}.
\end{equation}

Recent advances in CVQE have explored a wide range of deep learning architectures, reflecting various strategies to mitigate compression artifacts. Early approaches primarily relied on Convolutional Neural Networks (CNN) to perform intra-frame~\cite{qe-cnn, vrcnn, dncnn, irnn, dcad} enhancement by capturing local spatial features within individual frames, with ARCNN~\cite{arcnn} serving as a representative example.

With the emergence of attention mechanisms~\cite{attention}, attention-based methods~\cite{iirnet,ovqe,cpga,svkam,sta,stib} have gained traction for inter-frame enhancement by modeling long-range spatiotemporal dependencies, as exemplified by OVQE~\cite{ovqe}. By integrating a channel attention module into the residual FP and RD stages, OVQE achieved state-of-the-art performance and remains one of the highest-performing models to date.

More recently, hybrid architectures have been proposed, combining CNN with attention mechanisms~\cite{stff,m-swin,mdeformer,tvqe,ctve,stcf,mscaa} or incorporating additional components such as multi-layer perceptrons (MLP)~\cite{rivulet}. Notably, CTVE~\cite{ctve} exemplifies such modern hybrid designs through the integration of a U-Net~\cite{unet} backbone and Swin Transformer~\cite{swin-transformer} modules at the core of the model, offering enhanced robustness and adaptability across varying compression levels and diverse video content.

A comprehensive understanding of these evolving model architectures is crucial for developing CVQE solutions that generalize effectively across different compression standards.
\subsection{CVQE over standards}
\begin{figure}[!t]
    \centering
    \centerline{\includegraphics[width=0.45\textwidth]{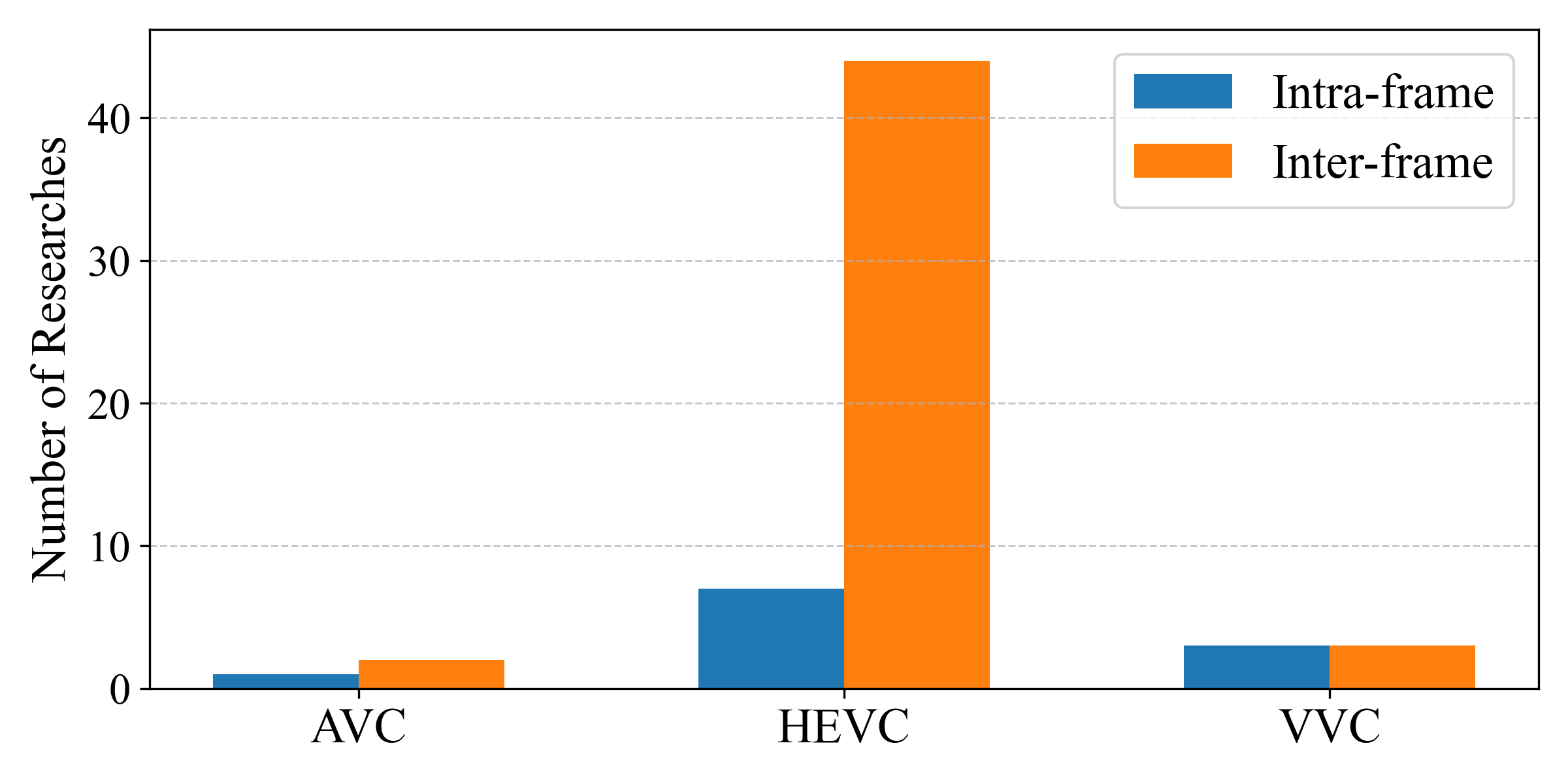}}
    \caption{Number of researches surveyed on CVQE.}
    \label{fig:survey_over_codec}
\vspace{-0.5cm}
\end{figure}
Compression standards such as H.264/AVC~\cite{avc-overview}, H.265/HEVC~\cite{hevc-overview}, and H.266/VVC~\cite{vvc-overview} are widely adopted in modern video transmission and storage systems, but they inevitably introduce compression artifacts that reduce visual quality, especially at lower bitrates. In research on CVQE, understanding these standards is essential because each codec produces different types and levels of artifacts. Fig.\ref{fig:survey_over_codec} illustrates the distribution of published research papers across the three codecs, showing the number of studies dedicated to each compression standard. 
\subsubsection{CVQE for H.264/AVC}
H.264/AVC still remains one of the most widely used video compression standards~\cite{bitmovin2024report}. However, because it was released before the widespread adoption of deep learning, research on CVQE specifically targeting AVC artifacts is relatively limited. Recent works like CAEF~\cite{caef}, Metabit~\cite{metabit}  pioneered CNN-based approaches for artifact reduction, but overall, this period saw few learning-based studies dedicated to AVC.
\subsubsection{CVQE for H.265/HEVC}
As illustrated in Fig.~\ref{fig:survey_over_codec} and Table~\ref{tab:overview_table}, H.265/HEVC remains the dominant codec in CVQE research. With the rise of standardized evaluation protocols, HEVC-based CVQE has seen significant advancements across multiple dimensions, including both intra-frame~\cite{arcnn,gl-fusion,vrcnn, dcad,sefcnn} and inter-frame~\cite{ovqe,stlvqe,tgafnet,mfmn,emafa,stdn,rfda,mrdn,svkam,stib,mfqe} enhancement strategies. Moreover, the incorporation of modern deep learning architectures, particularly attention-based~\cite{ovqe,cpga,sta,stawfn,iirnet} and hybrid models~\cite{rivulet,stff,pixrevive,ctve, m-swin}, has further accelerated progress in this area. Notably, STFF~\cite{stff}, a recent state-of-the-art approach, employs a hybrid architecture tailored for inter-frame enhancement and demonstrates substantial performance gains over previous methods. These trends highlight the central role of HEVC in the current CVQE landscape and underscore the importance of exploring increasingly sophisticated models to address compression artifacts effectively.

\subsubsection{CVQE for H.266/VVC}
H.266/VVC is the most recent video compression standard, offering significant bitrate reductions compared to its predecessors. However, given its relatively recent adoption, research efforts and standardized benchmarks targeting CVQE for VVC remain limited~\cite{irnn,stenet,ilf-qe,ovqe-vvc,hgrdn, mfqe-vvc}. Recent studies such as \cite{ovqe-vvc} and \cite{mfqe-vvc} primarily focus on artifact characterization or the direct adaptation of enhancement models originally designed for HEVC. As a result, there has been limited investigation into VVC-specific learning-based strategies that fully exploit the unique coding tools and structures introduced in this new standard.

\subsection{Compressed information utilization}
Most current approaches focus solely on processing decoded video frames~\cite{ovqe,stff,ctve,mdeformer,stdf,stdn,stdr,mfqr,mfmn,fast-mfqe,emafa,mfqe}, neglecting rich information embedded in the compressed domain, including motion vectors, residual signals, and coding tree unit (CTU) structures. Leveraging features directly from the compressed data can significantly lower computational overhead while retaining critical compression-aware cues~\cite{cpga, caef}. A prominent model is CPGA~\cite{cpga}, which incorporates compression information such as motion vectors, residuals, and predictive frames as additional features, achieving notable performance in CVQE. Nevertheless, designing models that can effectively leverage these compressed-domain features without compromising spatial and temporal coherence remains a challenging and unresolved task.
\section{Benchmarking Results And Discussion}
\label{sec:benchmarking}
\subsection{Experimental Setup}
\subsubsection{Objective Functions}
During training, CVQE models employ various loss functions. Early studies predominantly used Mean Squared Error (MSE) loss~\cite{qe-cnn,mfmn,twit,arcnn,irnn,bqev}, while more recent approaches favor the Charbonnier loss~\cite{charbonnier} for its enhanced robustness and perceptual fidelity~\cite{ovqe,mdeformer,mscaa,stenet,svkam,m-swin,tgafnet,caef,rivulet,cpga,stlvqe}. Some methods strategically combine both losses to leverage their complementary benefits~\cite{tvqe}. For optimization, most CVQE models adopt the ADAM optimizer~\cite{adam} due to its efficiency and adaptability across diverse training setups.

\subsubsection{Datasets and Configurations}

\begin{figure}[t]
\centering
    \begin{subfigure}{0.32\linewidth}
        \centering
        \includegraphics[width=\linewidth]{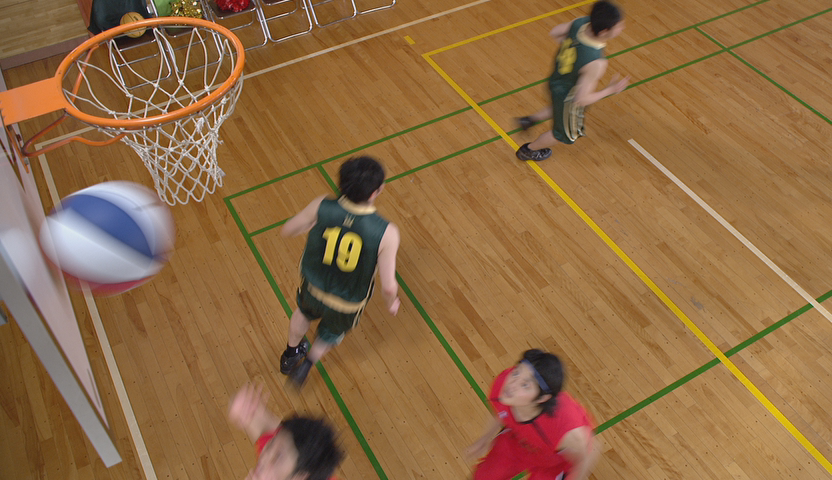}
        \caption{\centering BasketballDrill}
        \centering
    \end{subfigure}
    \begin{subfigure}{0.32\linewidth}
        \centering
        \includegraphics[width=\linewidth]{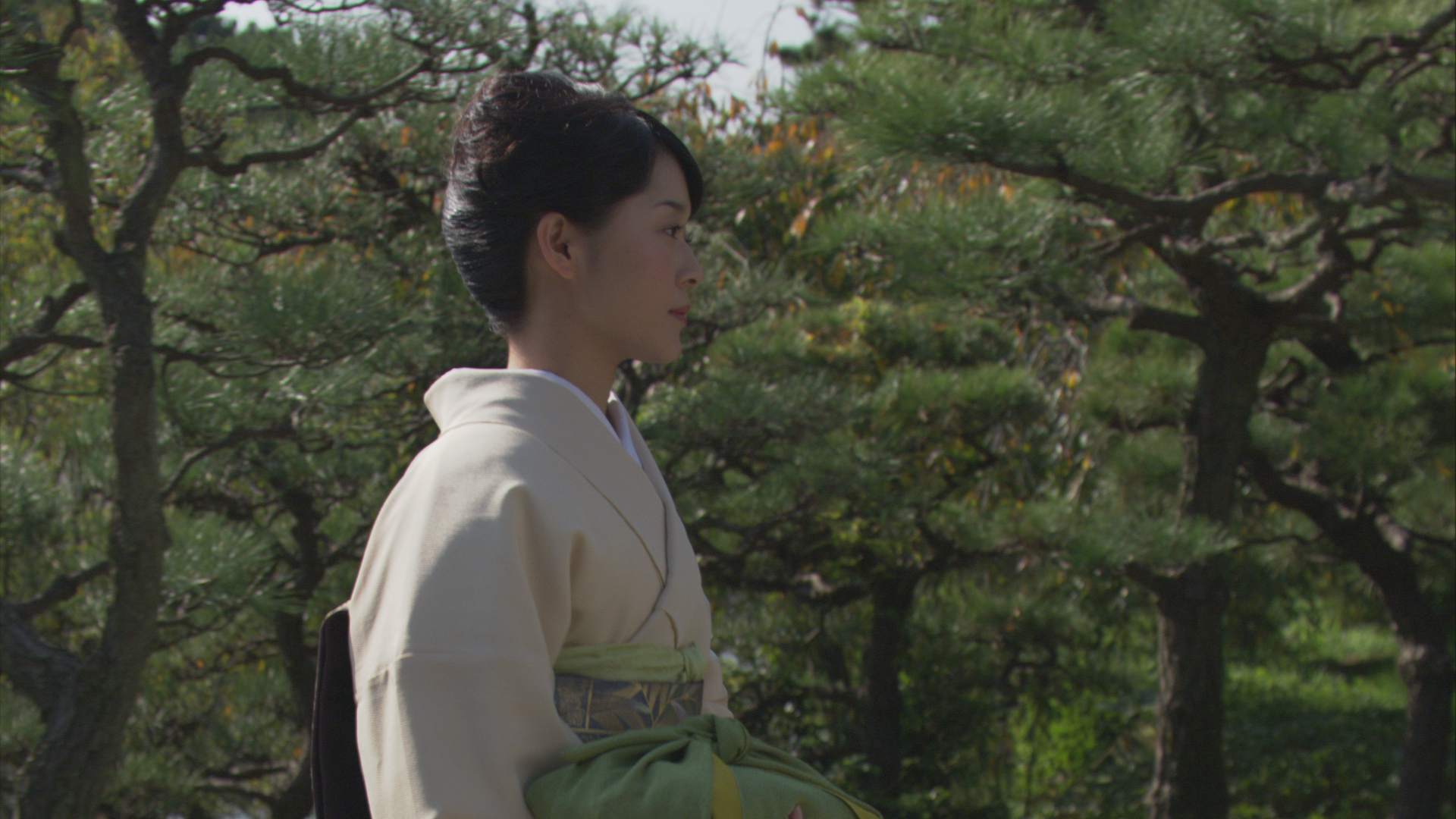}
        \caption{\centering Kimono1}
        \centering
    \end{subfigure}
    \begin{subfigure}{0.32\linewidth}
        \centering
        \includegraphics[width=\linewidth]{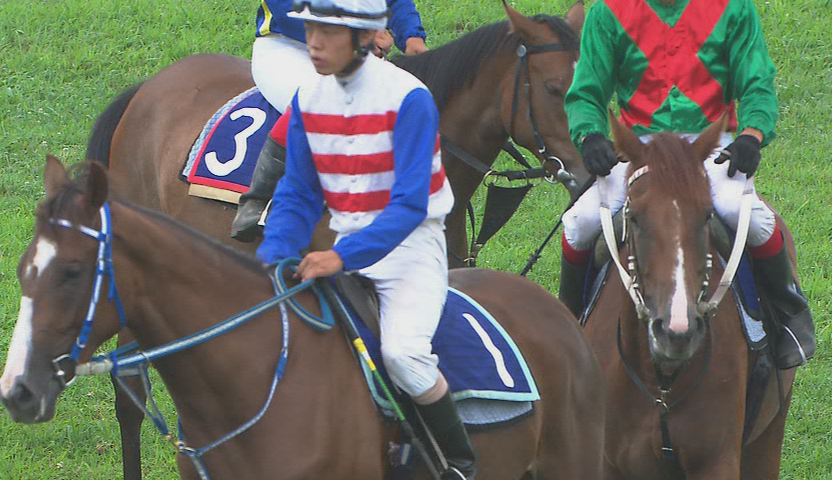}
        \caption{\centering RaceHorses}
        \centering
    \end{subfigure}
    \caption{The first frame of some test video sequences.}
    \label{fig:datatest}
\vspace{-0.5cm}
\end{figure}

For training CVQE models, MFQE dataset~\cite{mfqe}, which offers diverse low- and high-quality video pairs, is widely used~\cite{ovqe, mscaa, stff, mdeformer, stlvqe, m-swin, stawfn, mfqe}. To address the memory issue, training videos are randomly cropped to 128$\times$128~\cite{mdeformer, tvqe} or 64$\times$64~\cite{stff, ovqe}. Meanwhile, for evaluation, the JCT-VC~\cite{jctvc} test sequences offer a uniform benchmark set comprising 18 raw videos across five resolutions: 2560$\times$1600 (Class A), 1920$\times$1080 (Class B), 832$\times$480 (Class C), 416$\times$240 (Class D), and 1280$\times$720 (Class E). Fig.~\ref{fig:datatest} presents the first frame of several sequences (e.g., BasketballDrill, Kimono1, RaceHorses).


To ensure reproducible evaluations, each codec is tested with appropriate standard settings. H.264/AVC typically uses the CRF mode, while H.265/HEVC follows the CTC (Common Test Conditions)~\cite{ctc} using HM16.5 with the LDP configuration~\cite{mfqe2.0} to compress the test videos with five common quantization parameters (QP =  22, 27, 32, 37, and 42). For H.266/VVC, the VVenC encoder~\cite{vvenc} is commonly employed with standardized QP values.
\subsubsection{Evaluation Metrics}
To comprehensively assess the performance of CVQE methods, two main categories of evaluation metrics are typically considered. Quality improvement is evaluated by tracking changes in Peak Signal-to-Noise Ratio ($\Delta$PSNR) and Structural Similarity Index ($\Delta$SSIM), which reflect enhancements in visual fidelity and structural integrity relative to baseline codec outputs:
\begin{equation}
\label{eq:delta_quality}
\Delta Q = \frac{\sum_{t=1}^{n}Q_{t}^{HQ}-Q_{t}^{LQ}}{n},
\end{equation}
where $Q$ corresponds to the quality metric (e.g., PSNR, SSIM), $Q_{t}^{LQ}$ and $Q_{t}^{HQ}$ are the quality of the low- and high-quality reconstructed video frames at time index $t$, and $n$ is the number of frames used for evaluation.
In terms of computational complexity, Parameters and Floating Point Operations (FLOPs) are used to reflect processing efficiency and indicate the suitability of a method for real-time or resource-constrained applications. Together, these metrics provide a comprehensive evaluation of visual quality and computational practicality.
\subsection{Results and Discussion}
\begin{figure}[!t]
    \centering
    \centerline{\includegraphics[width=0.45\textwidth]{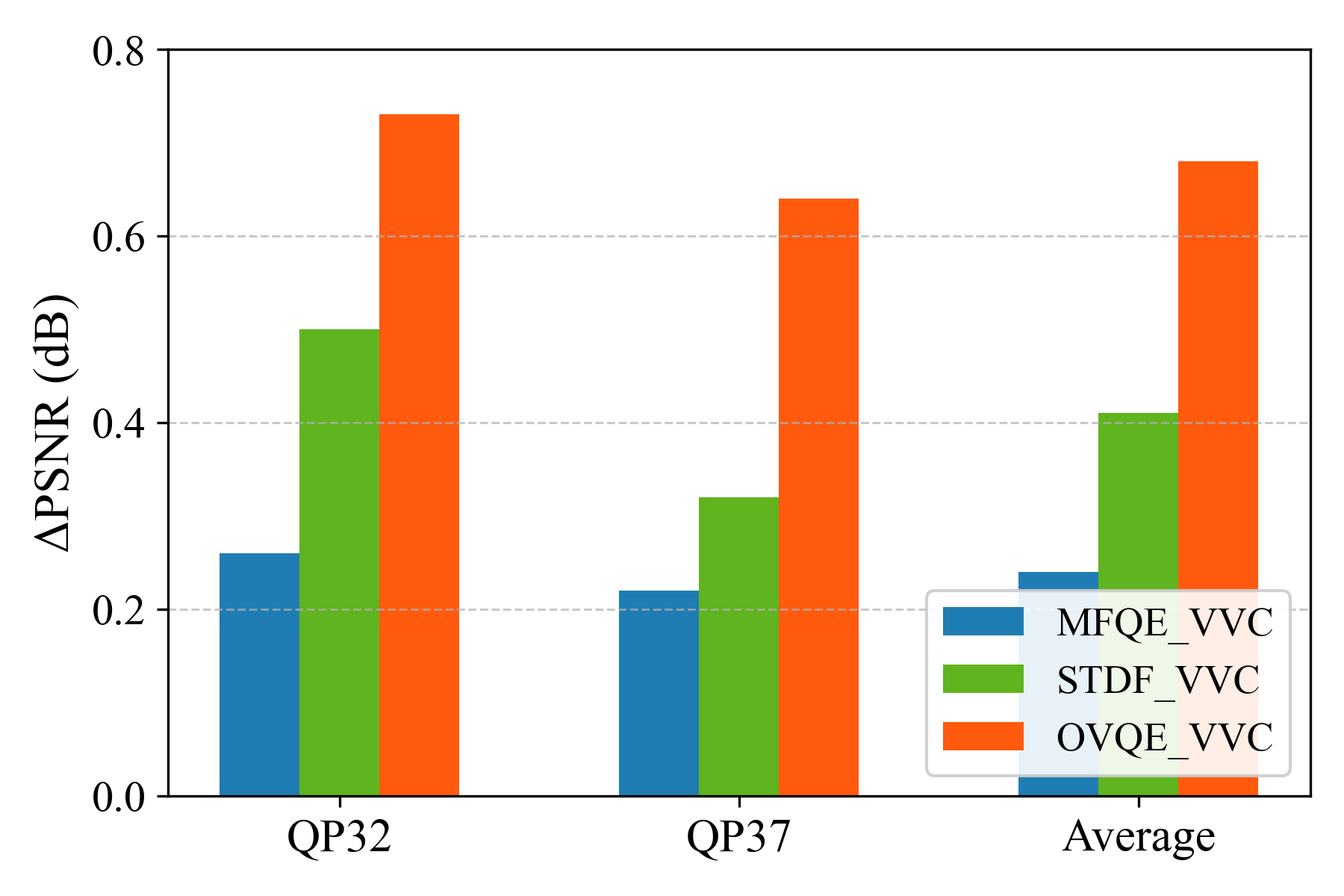}}
    \vspace{-1em}
    \caption{Performance comparison of CVQE methods with H.266/VVC compressed videos using metrics of  $\Delta\mathrm{PSNR}\ (\mathrm{dB})$.}
    \label{fig:VVC_results}
\end{figure}
\begin{figure}[!t]
    \centering
    \centerline{\includegraphics[width=0.5\textwidth]{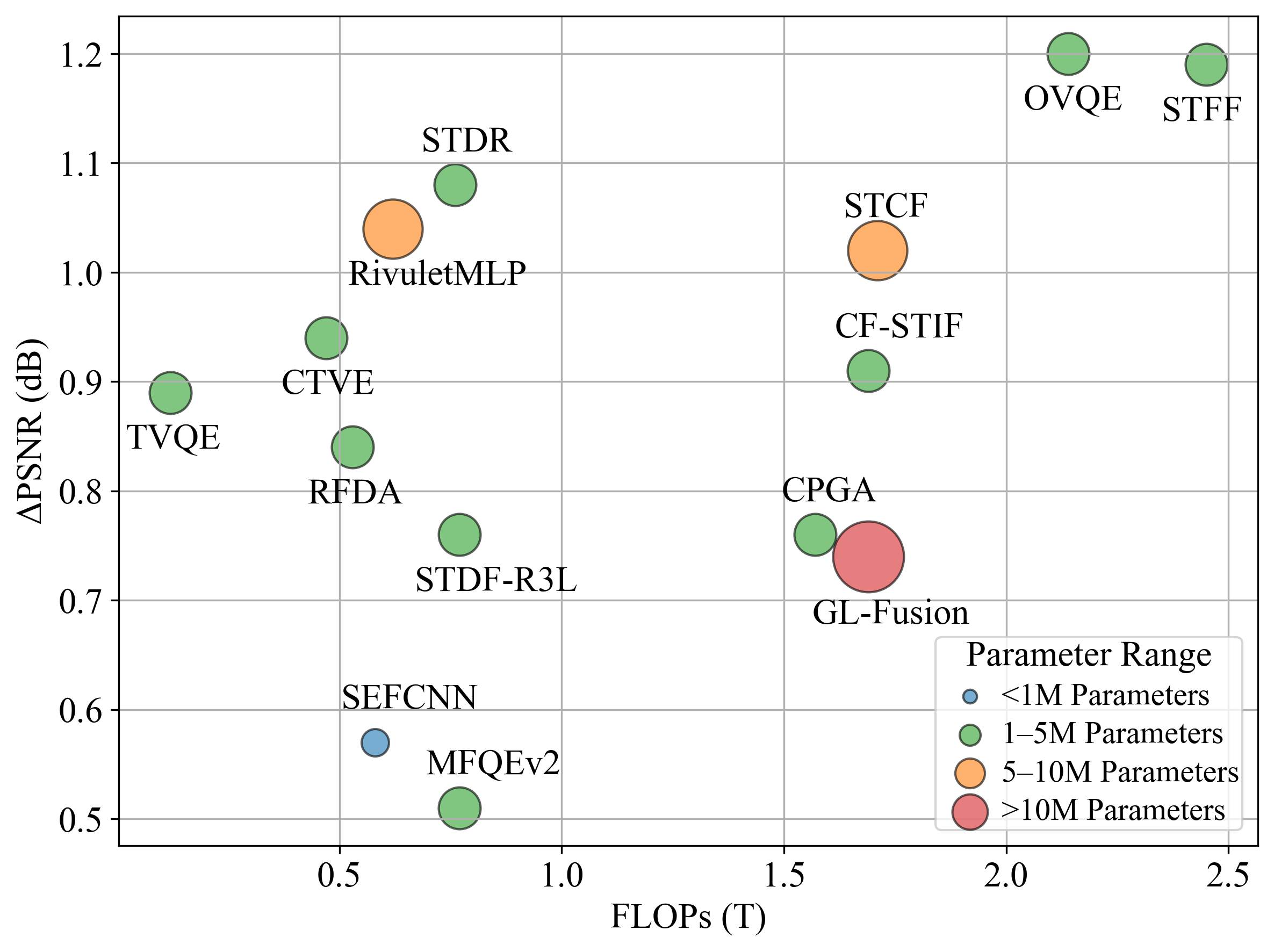}}
    \caption{Trade-offs between $\Delta\mathrm{PSNR}\ (\mathrm{dB})$ improvement, FLOPs, and number of Parameters for various CVQE models.}
    \vspace{-1em}
    \label{fig:PSNR_Params_FLOPs}
\end{figure}
\begin{table}[t]
\centering
\scriptsize
\begin{threeparttable}
\caption{Performance comparison of CVQE methods with H.264/AVC compressed videos using metrics of  $\Delta\mathrm{PSNR}\ (\mathrm{dB}) \,/\, \Delta\mathrm{SSIM}\ (\times 10^{-2})$.}
\label{tab:avc_result}
\newcolumntype{R}{>{\raggedleft\arraybackslash}X}
\newcolumntype{C}{>{\centering\arraybackslash}X}
\begin{tabularx}{\columnwidth}{lCC} 

\hline
Year           & 2024        & 2025        \\ \hline
Model          & MetaBit~\cite{metabit}     & CAEF~\cite{caef}        \\
Coding Type    & Inter-frame & Inter-frame \\ \hline
CRF15          & N/A / N/A       & 1.20 / 0.57 \\
CRF25          & N/A / N/A       & 0.99 / 1.05 \\
CRF35          & 0.96 / 2.30 & 0.94 / 2.34 \\
CRF40          & 1.03 / 3.10 & N/A / N/A       \\
CRF50          & 0.88 / 4.20 & N/A / N/A       \\
Average        & 0.96 / 3.20 & 1.04 / 1.32 \\ \hline
Parameters (M) & N/A           & 4.56        \\ \hline

\end{tabularx}
\vspace{0.1cm}
\scriptsize 
\end{threeparttable}
\vspace{-0.5cm}
\end{table}
\begin{table*}[t]
\centering
\scriptsize
\begin{threeparttable}
\caption{Performance comparison of CVQE methods with H.265/HEVC compressed videos using metrics of  $\Delta\mathrm{PSNR}\ (\mathrm{dB}) \,/\, \Delta\mathrm{SSIM}\ (\times 10^{-2})$.}
\label{tab:hevc_result}
\setlength{\dashlinedash}{1pt}    
\setlength{\dashlinegap}{1pt}     
\setlength{\arrayrulewidth}{0.3pt} 
\begin{tabular}{cclccccccc}
\hline
Coding Type                   & Year                    & Model                      & QP37                                                            & QP32                                                            & QP27                                                            & QP22                                                            & Average                                                         & Parameters (M)                 & FLOPs (T)                       \\ \hline
\multirow{31}{*}{Inter-frame} & \multirow{6}{*}{2025}   & MSCAA \cite{mscaa}         & 1.02 / 1.80                                                     & 0.98 / 1.28                                                     & 0.92 / 0.81                                                     & 0.83 / 0.51                                                     & 0.94 / 1.10                                                     & 1.71                           & N/A                             \\
                              &                         & STFF \cite{stff}           & \textcolor{red}{\textbf{1.20}} / \textcolor{red}{\textbf{2.20}} & \textcolor{red}{\textbf{1.24}} / \textcolor{blue}{\underline{1.58}} & \textcolor{blue}{\underline{1.24}} / 1.04                           & \textcolor{blue}{\underline{1.09}} / \textcolor{blue}{\underline{0.60}} & \textcolor{blue}{\underline{1.19}} / \textcolor{blue}{\underline{1.36}} & 4.45                           & 2.45                            \\
                              &                         & MFMN \cite{mfmn}           & 0.44 / 1.01                                                     & 0.39 / 0.66                                                     & 0.35 / 0.45                                                     & 0.26 / 0.25                                                     & 0.36 / 0.59                                                     & N/A                            & \textbf{\textcolor{red}{0.06}}  \\
                              &                         & IIRNet \cite{iirnet}       & 1.01 / 1.81                                                     & 0.96 / 1.24                                                     & 0.95 / 0.84                                                     & 0.89 / 0.50                                                     & 0.95 / 1.10                                                     & N/A                            & N/A                             \\
                              &                         & RivuletMLP \cite{rivulet}  & 1.13 / 1.93                                                     & 1.05 / 1.37                                                     & 0.95 / 0.90                                                     & N/A / N/A                                                       & 1.04 / 1.40                                                     & 6.90                           & 0.62                            \\
                              &                         & MDEformer \cite{mdeformer} & 0.99 / 1.79                                                     & 1.03 / 1.25                                                     & 1.05 / 0.84                                                     & 0.93 / 0.52                                                     & 1.00 / 1.10                                                     & N/A                            & N/A                             \\ \cdashline{2-10}
                              & \multirow{10}{*}{2024~} & TVQE \cite{tvqe}           & 0.98 / 1.82                                                     & 0.93 / 1.24                                                     & 0.87 / 0.80                                                     & 0.77 / 0.49                                                     & 0.89 / 1.09                                                     & 2.09                           & 0.12                            \\
                              &                         & CTVE \cite{ctve}           & 0.96 / 1.78                                                     & 0.96 / 1.28                                                     & 0.94 / 0.82                                                     & 0.88 / 0.50                                                     & 0.94 / 1.10                                                     & 2.96                           & 0.47                            \\
                              &                         & M-Swin \cite{m-swin}       & 1.04 / 1.91                                                     & 1.06 / 1.35                                                     & 1.00 / 0.85                                                     & 0.89 / 0.49                                                     & 1.00 / 1.15                                                     & 1.62                           & N/A                             \\
                              &                         & CPGA~\cite{cpga}           & 0.82 / 1.56                                                     & 0.83 / 1.10                                                     & 0.77 / 0.75                                                     & 0.62 / 0.39                                                     & 0.76 / 0.95                                                     & 1.39                           & 1.57                            \\
                              &                         & EMAFA~\cite{emafa}         & 1.06 / 1.89                                                     & \textcolor{blue}{\underline{1.12} }/ 1.40                           & 1.10 / 0.90                                                     & 0.95 / 0.54                                                     & 1.06 / 1.18                                                     & 3.55                           & N/A                             \\
                              &                         & SVKAM \cite{svkam}         & 1.09 / 1.95                                                     & 0.96 / 1.46                                                     & 0.90 / 0.94                                                     & 0.85 / 0.65                                                     & 0.95 / 1.25                                                     & 3.53                           & N/A                             \\
                              &                         & TGAFNet \cite{tgafnet}     & 0.96 / 1.77                                                     & 1.00 / 1.26                                                     & 0.99 / 0.83                                                     & 0.90 / 0.50                                                     & 0.96 / 1.09                                                     & 1.40                           & N/A                             \\
                              &                         & BQEV \cite{bqev}           & 0.65 / N/A                                                      & 0.56 / N/A                                                      & 0.46 / N/A                                                      & 0.31 / N/A                                                      & 0.50 / N/A                                                      & N/A                            & N/A                             \\
                              &                         & STAWFN \cite{stawfn}       & 1.02 / 1.78                                                     & 1.02 / 1.24                                                     & 1.01 / 0.84                                                     & 0.90 / 0.51                                                     & 0.99 / 1.09                                                     & 1.55                           & N/A                             \\
                              &                         & PixRevive \cite{pixrevive} & 1.08 / 1.93                                                     & 1.09 / 1.55                                                     & 1.03 / \textcolor{red}{\textbf{1.17}}                           & N/A / N/A                                                       & 1.07 / 1.55                                                     & N/A                            & N/A                             \\ \cdashline{2-10}
                              & \multirow{7}{*}{2023}   & Fast-MFQE \cite{fast-mfqe} & 0.67 / 1.18                                                     & 0.63 / 0.69                                                     & 0.55 / 0.48                                                     & N/A / N/A                                                       & 0.62 / 0.78                                                     & \textbf{\textcolor{red}{0.24}} & N/A                             \\
                              &                         & OVQE \cite{ovqe}           & \textcolor{blue}{\underline{1.17} }/ \textcolor{blue}{\underline{2.14}} & \textcolor{red}{\textbf{1.24 }}/ \textcolor{red}{\textbf{2.74}} & \textcolor{red}{\textbf{1.25 }}/ \textcolor{blue}{\underline{1.07}} & \textcolor{red}{\textbf{1.12 }}/ \textcolor{red}{\textbf{0.63}} & \textcolor{red}{\textbf{1.20 }}/ \textcolor{red}{\textbf{1.65}} & 3.11                           & 2.14                            \\
                              &                         & STCF \cite{stcf}           & 1.02 / 1.81                                                     & 1.07 / 1.32                                                     & 1.05 / 0.88                                                     & 0.93 / 0.54                                                     & 1.02 / 1.14                                                     & 7.07                           & 1.71                            \\
                              &                         & STDN \cite{stdn}           & 0.95 / 1.68                                                     & 0.98 / 1.20                                                     & 0.97 / 0.80                                                     & 0.85 / 0.48                                                     & 0.94 / 1.04                                                     & N/A                            & N/A                             \\
                              &                         & STDR \cite{stdr}           & 0.98 / 1.79                                                     & 1.24 / 1.24                                                     & 0.81 / 0.81                                                     & 0.48 / 0.48                                                     & 1.08 / 1.08                                                     & 1.32                           & 0.76                            \\
                              &                         & PIMnet \cite{pimnet}       & 0.97 / 1.74                                                     & 1.06 / 1.33                                                     & 1.05 / 0.88                                                     & 0.97 / 0.53                                                     & 1.01 / 1.12                                                     & 3.89                           & N/A                             \\
                              &                         & STIB \cite{stib}           & 0.95 / 1.74                                                     & 0.96 / 1.22                                                     & 0.87 / 0.73                                                     & 0.84 / 0.47                                                     & 0.91 / 1.04                                                     & 1.42                           & N/A                             \\ \cdashline{2-10}
                              & \multirow{4}{*}{2022}   & FastCNN \cite{fastcnn}     & 0.85 / 1.51                                                     & 0.83 / 1.03                                                     & 0.76 / 0.65                                                     & 0.67 / 0.39                                                     & 0.78 / 0.90                                                     & \underline{\textcolor{blue}{0.43}} & N/A                             \\
                              &                         & MRDN \cite{mrdn}           & 0.78 / 1.47                                                     & 0.81 / 1.02                                                     & 0.83 / 0.72                                                     & 0.70 / 0.40                                                     & 0.78 / 0.90                                                     & N/A                            & N/A                             \\
                              &                         & CF-STIF \cite{cf-stif}     & 0.92 / 1.67                                                     & 0.95 / 1.18                                                     & 0.94 / 0.78                                                     & 0.85 / 0.46                                                     & 0.91 / 1.02                                                     & 2.22                           & 1.69                            \\
                              &                         & EAAGA \cite{eaaga}         & 1.04 / 1.86                                                     & 1.01 / 1.25                                                     & 1.00 / 0.86                                                     & 0.81 / 0.48                                                     & 0.97 / 1.11                                                     & N/A                            & N/A                             \\ \cdashline{2-10}
                              & 2021                    & RFDA \cite{rfda}           & 0.91 / 1.62                                                     & 0.87 / 1.07                                                     & 0.82 / 0.68                                                     & 0.76 / 0.42                                                     & 0.84 / 0.95                                                     & 1.27                           & \underline{\textcolor{blue}{0.53}}  \\ \cdashline{2-10}
                              & 2020                    & STDF-R3L \cite{stdf}       & 0.83 / 1.51                                                     & 0.86 / 1.04                                                     & 0.72 / 0.57                                                     & 0.63 / 0.34                                                     & 0.76 / 0.87                                                     & 1.28                           & 0.77                            \\ \cdashline{2-10}
                              & 2019                    & MFQEv2 \cite{mfqe2.0}      & 0.56 / 1.09                                                     & 0.52 / 0.68                                                     & 0.49 / 0.42                                                     & 0.46 / 0.27                                                     & 0.51 / 0.62                                                     & 1.62                           & 0.77                            \\ \cdashline{2-10}
                              & 2018                    & MFQE \cite{mfqe}           & 0.46 / 0.88                                                     & 0.43 / 0.58                                                     & 0.40 / 0.34                                                     & 0.31 / 0.19                                                     & 0.40 / 0.50                                                     & 1.79                           & N/A                             \\ \hline
\multicolumn{1}{l}{}          &                         &                            & \multicolumn{1}{l}{}                                            & \multicolumn{1}{l}{}                                            & \multicolumn{1}{l}{}                                            & \multicolumn{1}{l}{}                                            & \multicolumn{1}{l}{}                                            & \multicolumn{1}{l}{}           & \multicolumn{1}{l}{}            \\ \hline
\multirow{7}{*}{Intra-frame}  & \multirow{2}{*}{2022}                    & SEFCNN \cite{sefcnn}       & \textcolor{blue}{\underline{0.68} / }N/A                            & \textcolor{blue}{\underline{0.63}} / N/A                            & \textcolor{blue}{\underline{0.49}} / N/A                            & \textcolor{blue}{\underline{0.46}} / N/A                            & \textcolor{blue}{\underline{0.57} }/ N/A                            & 0.64                           & 0.58                            \\ 
                              &                     & GL-Fusion \cite{gl-fusion} & \textcolor{red}{\textbf{0.89} }/ N/A                            & \textcolor{red}{\textbf{0.82 }}/ N/A                            & \textbf{\textcolor{red}{0.68} }/ N/A                            & \textbf{\textcolor{red}{0.55} }/ N/A                            & \textcolor{red}{\textbf{0.74}} / N/A                            & 35.53                          & 1.69                            \\ \cdashline{2-10}
                              & 2020                    & RNAN \cite{rnan}           & 0.44 / \textbf{\textcolor{red}{0.95}}                           & 0.41 / \textbf{\textcolor{red}{0.62}}                           & N/A / N/A                                                       & N/A / N/A                                                       & 0.43 / 0.79                                                     & 7.41                           & N/A                             \\ \cdashline{2-10}
                              & 2017                    & DnCNN \cite{dncnn}         & 0.26 / \underline{\textcolor{blue}{0.58}}                           & 0.26 / 0.35                                                     & 0.27 / \underline{\textcolor{blue}{0.24}}                           & 0.29 / \underline{\textcolor{blue}{0.18}}                           & 0.27 / \underline{\textcolor{blue}{0.34}}                           & N/A                            & N/A                             \\ \cdashline{2-10}
                              & \multirow{2}{*}{2016}   & VRCNN \cite{vrcnn}         & 0.32 / 0.35                                                     & 0.28 / 0.15                                                     & 0.24 / 0.09                                                     & N/A / N/A                                                       & 0.23 / 0.20                                                     & \textbf{\textcolor{red}{0.05}} & 0.05                            \\
                              &                         & DCAD \cite{dcad}           & 0.32 / 0.67                                                     & 0.32 / \underline{\textcolor{blue}{0.44}}                           & 0.32 / \textbf{\textcolor{red}{0.30}}                           & 0.31 / \textbf{\textcolor{red}{0.19}}                           & 0.32 / \textbf{\textcolor{red}{0.40}}                           & N/A                            & N/A                             \\ \cdashline{2-10}
                              & 2015                    & AR-CNN \cite{arcnn}        & 0.23 / 0.45                                                     & 0.18 / 0.19                                                     & 0.18 / 0.14                                                     & 0.14 / 0.08                                                     & 0.18 / 0.22                                                     & \underline{\textcolor{blue}{0.11}} & \underline{\textcolor{blue}{0.10}}  \\ \hline
\end{tabular}
\scriptsize
* \textcolor{red}{\textbf{Red}} indicates the best result, \textcolor{blue}{\underline{Blue}} indicates the second best result within each column. \\
* FLOPs is measured with the input resolution of 720p.
\end{threeparttable}
\vspace{-0.5cm}
\end{table*}


Table~\ref{tab:avc_result} presents the performance of representative inter-frame CVQE models on H.264/AVC compressed videos. Despite limited benchmarking data across all compression levels, the results reflect a consistent trend: recent inter-frame approaches, such as MetaBit~\cite{metabit} and CAEF~\cite{caef} (which additionally leverage compressed-domain information), demonstrate promising enhancement capabilities.

Fig.~\ref{fig:VVC_results} underscores the nascent state of VVC quality enhancement research, reflecting the standard's recent adoption. Current approaches predominantly adapt existing HEVC-focused models (e.g., OVQE\_VVC~\cite{ovqe-vvc}, MFQE\_VVC~\cite{mfqe-vvc}, STDF\_VVC) rather than designing VVC-specific solutions. Notably, OVQE~\cite{ovqe}, originally designed for HEVC, still achieves the best performance on VVC sequences, with an average gain of 0.68 dB. These findings suggest that while initial adaptations are effective, VVC-specific designs remain necessary.

Table~\ref{tab:hevc_result} and Fig.~\ref{fig:PSNR_Params_FLOPs}, which summarize results from multiple HEVC enhancement studies across four QP levels, highlight several key observations:
\begin{itemize}
    \item \textbf{OVQE}~\cite{ovqe} model archives the best performance in both $\Delta$PSNR and $\Delta$SSIM indexes with 1.20 dB and 1.65, followed by STFF~\cite{stff} with the results of 1.19 dB and 1.36 respectively.
    \item \textbf{Attention-based models} achieve the highest average $\Delta$PSNR, with OVQE reaching 1.20 dB, approximately 0.80 dB higher than early CNN-based methods such as MFQE~\cite{mfqe} (0.40 dB). However, this performance gain comes at the cost of high computational complexity (2.14 TFLOPs), which limits their suitability for real-time deployment.
    \item \textbf{Hybrid architectures} demonstrate favorable trade-offs between performance and efficiency. CTVE~\cite{ctve} achieves near-transformer reconstruction quality (0.94 dB $\Delta\mathrm{PSNR}$) while reducing computational load by 78\% compared to OVQE (0.47 TFLOPs vs. 2.14 TFLOPs). Meanwhile, STFF~\cite{stff} offers superior perceptual quality (1.19 dB) with 42\% fewer parameters than comparable pure attention-based models, such as STCF~\cite{stcf} (4.45M vs. 7.07M).
    \item \textbf{Intra-frame methods} do not deliver competitive performance compared to \textbf{inter-frame} approaches. The best-performing intra-frame model, GL-Fusion~\cite{gl-fusion}, achieves 0.74 dB, still falling significantly behind the best inter-frame counterpart, OVQE, by a margin of 0.46 dB.
\end{itemize} 

\section{Conclusions}
This paper presents a comprehensive survey of learning-based compressed video quality enhancement, addressing gaps in the existing literature through a novel classification framework that systematically categorizes CVQE approaches across architectural paradigms, video coding standards, and compressed domain information utilization. We propose a comprehensive benchmarking protocol that integrates modern codecs (H.264/AVC, H.265/HEVC, H.266/VVC) with consistent test sequences and multi-criteria metrics. Through extensive comparative analysis, we highlighted key trade-offs between reconstruction performance and computational complexity, emphasizing the strengths of hybrid models in balancing quality and efficiency. Additionally, our survey shows that while inter-frame and attention-based methods achieve superior performance, real-time deployment challenges remain due to high computational demands, indicating that future research should focus on VVC-compressed video quality enhancement.
\label{sec:conclusions}

\bibliographystyle{IEEEtran}
\bibliography{IEEEabrv,paper}

\end{document}